\documentclass{article}

\usepackage{arxiv}

\usepackage[utf8]{inputenc} 
\usepackage[T1]{fontenc}    
\usepackage{hyperref}       
\usepackage{url}            
\usepackage{booktabs}       
\usepackage{amsfonts}       
\usepackage{nicefrac}       
\usepackage{microtype}      
\usepackage{lipsum}
\usepackage{tcolorbox}
\usepackage{graphicx}
\graphicspath{ {./images/} }

\newcommand{\al}{\textit{et al.}}

\title{Towards Automated Acceptance testing for industrial robots}

\author{
Marcela G. dos Santos \\
SmArtSE Research Team \\
Université du Québec à Chicoutimi \\
555, boulevard de l’Université, G7H 2B1 \\
Chicoutimi, QC, Canada \\
\texttt{marcela.santos1@uqac.ca} \\
\And

Fabio Petrillo \\
SmArtSE Research Team \\
Université du Québec à Chicoutimi \\
555, boulevard de l’Université, G7H 2B1 \\
Chicoutimi, QC, Canada \\ 
\texttt{fabio@petrillo.com} \\
}

\begin{document}
\maketitle

\begin{abstract}
 Industrial robots are important machines applied in numerous modern industries that execute repetitive tasks with high accuracy, replacing or supporting dangerous jobs. In this kind of system, with increased complexity in which cost is related to the time the system keeps working, the system must operate with a minimum number of failures. In other words, a quality aspect important in industry is reliability.  We hypothesize that Automated Acceptance Testing improves reliability for industrial robot program. We present the research question, the motivation for this study, our hypothesis and future research efforts.
\end{abstract}


\section{Introduction}
\textbf{Industrial robots} execute repetitive tasks (as picking-and-placing, soldering, and cutting) with high accuracy, safety (replacing or supporting people in dangerous jobs) and efficiency in numerous industries like automotive,  electrical/electronics, metal, and machinery \cite{IFR2019}. Hence, the number of industrial robots increased by +400K shipped units globally in 2018, with sales of  US\$ 16.5 billion \cite{IFR2019}. Consequently, the industrial robot's marketing is forecast to achieve 66.48 billion by 2027 \cite{FORTUNE2019}.
 
According to \cite{GARCIA2020} one of top main challenges in robotics is achieving robustness. Robustness is defined as the ability of the system to recover from errors and continue operating with reliability for a period of time. This definition lead us from the robustness to reliability. Reliability is the probability of the software  operates without failure for a specific amount of time, in a specific environment for a purpose \cite{SQQA2018}. So, in order to guarantee robustness, we first need to guarantee \textbf{reliability} in a software. Our research question is \textit{How can we improve reliability in the industrial robots?}



Software reliability is a quality aspect that determines the degree with which a system operates free of failures over a specified time, in a given environment, for a specific purpose \cite{SOMMERVILLE2015}, and an important facet of reliability is fault-tolerance, the degree with which a system operates as intended despite the presence of hardware or software faults \cite{ISO25010}. A known software failure that affect robotic systems is when NASA's Curiosity rover, a mobile robot that explores Mars, eventually found itself in software trouble \cite{ATKINSON2020}, it loses its sense of direction and stopped en route due to a software bug. Another example of software fail a complex system is the two flights crash happened due to faulty software in the Boeing 737 MAX maneuver system, killing hundreds of people \cite{BOEINGCRASH1}.

Most of these failures originate from undetected system faults, and, in practice, it is difficult to guarantee failure-free software \cite{SOMMERVILLE2015}. To reduce failures and \textbf{overcome software reliability challenges}, Garcia \al propose rigorous development and \textbf{test processes} to perform solutions \cite{GARCIA2020}. 

However, there are quality \textbf{gaps related to software development in the robotics industry} that focuses, in most cases, on performance and demonstration of certain functionalities \cite{CHUNG2007}. In the empirical study to assess the state-of-the-art and practice of robotics software engineering performed by Garcia \al \cite{GARCIA2020}, they highlight the difficulty and the importance of programming robots to handle failures, as well as systematic testing processes can support the management of possible failures and unexpected environmental events over prolonged time frames.

Some studies address testing for robotic systems \cite{ASHRAF2020, BRETL2008, ESTIVILLCASTRO2018, MOSSIGE2015, CHUNG2007, ERICH2019}, but to the best of our knowledge, this is the first work that shows the application of automated acceptance for industrial robots using simulators. We propose an infrastructure of automated Acceptance Testing using simulators to improve reliability for industrial robots. This paper's main contribution is to present our hypothesis and our detailing plan to future research efforts.


This paper's main contribution is to present our hypothesis and our detailing plan to future research efforts. Our study is organized as follows. Section \ref{sec:BACKGROUND} defines the concepts of industrial robots programming and acceptance testing. Section \ref{sec:PROPOSAL} describes in detail our proposal. Section \ref{sec:RELATEDWORK}  shows related studies. Section \ref{sec:CONCLUSIONS} synthesizes the final remarks and future work.

\section{Background}
\label{sec:BACKGROUND}

\begin{figure}[htbp]
\centering
 \includegraphics[width=.55\linewidth]{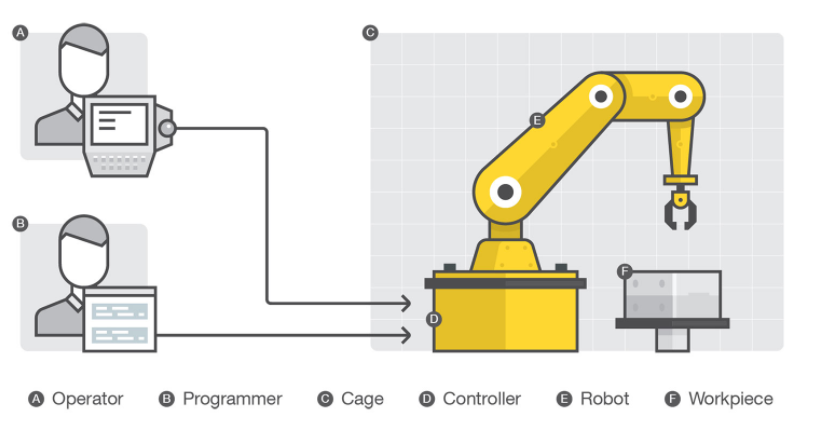}
\caption{Industrial robot programming \cite{TRENDMICRO2021}}.
\label{fig:overviewrobotindustrialprogramming}
\end{figure}

Figure \ref{fig:overviewrobotindustrialprogramming} shows an industrial robot on the shop floor. 
Industrial robotic systems are complex hardware-software. The hardware (E) is in a cage (C) that is a structure used to protect the human from the robot's incorrect functionality.


The software plays a critical role in the IRS, and it has two parts.  A controller (D) is responsible for translating the commands so that the industrial robot understands and executes them. The \textbf{robot program}, a script software component, defines this desired behaviour \cite{AHMAD2016}. 

The software plays a critical role in the IRS, and it has two parts.  A controller is responsible for translating the commands so that the industrial robot understands and executes them. The \textbf{robot program}, a script software component, defines this desired behaviour \cite{AHMAD2016}. 

The robot program is crucial concerning the business that uses IRS. With it, it is possible to adequate robot's behaviour according to the business requirements and play different tasks. Thus, we can write the robot program and send commands for the industrial robots via the operator (A) or programmer (B).

The robot program is crucial concerning the business that uses IRS. With it, it is possible to adequate robot's behaviour according to the business requirements and play different tasks. Thus, we can write the robot program and send commands for the industrial robots via the operator or programmer.

The industrial robot programming methods are online and off-line programming. In the online method, the operator programs the robot on the shop floor, and the industry needs to stop the production. There are two sub-methods in online programming. Lead through is when the operator takes the robot manually and guides it through the trajectory.

The other way to program an industrial robot in the method online is using the teach-pendant. First, the operator guides the manipulator to specific points. After that, the operator records these points to composed a trajectory. Finally, the manipulator executes the trajectory. To record another trajectory, the operator needs to take off the robot, and the operator needs to teach the points using the teach-pendant.

Whereas to program the robot in the online method, the operator needs to stop the production line, in the off-line method, the robot can be programming without interruption on the shop floor. Off-line programming is a method in which the robot is programming in an environment composed of Industrial Robot Languages (IRPLs) and/or simulation software. IRPLs are purpose-built, domain-specific programming languages that include special instructions to move the robot’s arm(s), standard control-flow instructions and APIs to access low-level resources \cite{POGLIANI2020}. 

There are two types of IRPLs and Simulators, the owners and generics simulators. For example, Kuka, one of the most popular industrial robot vendors, has its language, KRL and Kuka.Sim the simulation software for offline programming. Another example is the language RAPID, the simulator ABB Robot Studio for ABB, other industrial robot vendors. In the owner environment, the programming made is only for the brand's robots.On generic simulators, as Delfoi, Octobuz, RoboDK, RoboExpert, Delmia, Process Simulate and Robcad, it is possible to program different robots' brands. 

In both types of programming environment, it is possible to program directly in text or using the simulation workstation. The number of simulators has grown substantially in the last years as Ivaldi \al \cite{IVALDI2014} observed in them study. They performed a survey with 119 participants and observed one of the main purpose to use simulators is test software controller. 

 However, \textbf{simulation and robustness} are testing that are not well established yet as observed  Afzal \al \cite{AFZAL2020} in them study. They identify a total of \textbf{12 testing practices for robotic systems}:  field testing, logging and playback, simulation testing, plan-based testing, compliance testing, unit testing, performance testing, hardware testing, robustness testing, regression testing, continuous integration and test maintenance. One reason is the gap related supporting tools and infrastructure around them simulation and robustness testing.


Acceptance testing determines if the system satisfies the acceptance criteria (AC). AC is the criteria that ensure that the system quality is acceptable, and quality requirements determine the system quality. Some examples of quality requirements are functional correctness, completeness, accuracy, data integrity, usability, performance, start-up,  reliability, availability, maintainability and robustness \cite{SQQA2018}. So we can create acceptance criteria for each quality requirements, for example reliability. 

The role of acceptance testing is to help the customer to determine if the system functions in accordance with the customer’s expectations, and is written by the customer and not by the developers. Then, one of challenges is to define in which language. 

Programming languages are generally not well understood by the customer and natural language has some aspects that are not suitable to writing acceptance tests: too vague, too verbose and the complex grammar that structures written communication is largely unnecessary for testing. A good way is teamwork between the customers and the developers team to investigate sources of acceptance tests, and define common domain language to write acceptance testing and the utilization of acceptance testing frameworks \cite{ROGERS2004}. 

\begin{figure*}[htpb]
\centering
\includegraphics[width=\linewidth]{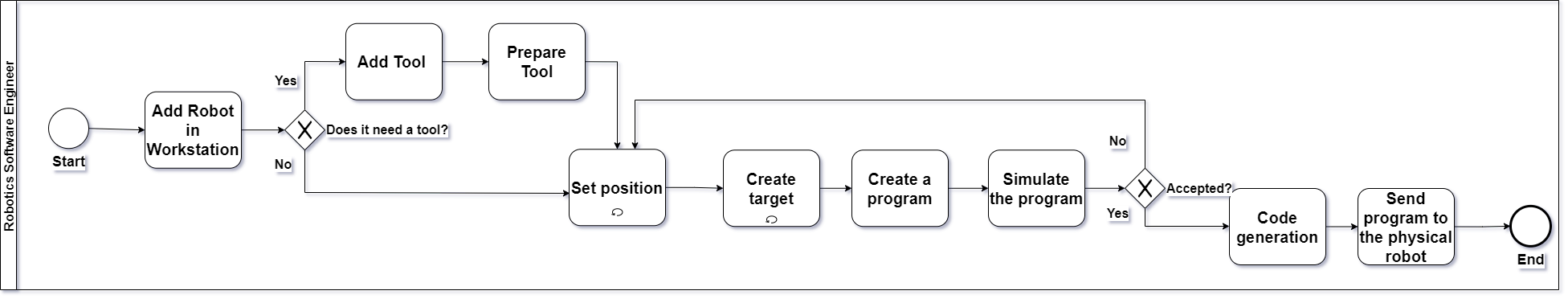}
\caption{Industrial robot programming using simulators.}
\label{fig:overviewrobotindustrialprogrammingbpmn}
\end{figure*}

Nowadays, there is one main role related to the robot programming in industrial environment. The role is responsible for writing the robot program in simulator, verifying if the robot has a behaviour correct, and finally sending the program generated by the simulator to the industrial robot on the shop floor. Our study uses Robotics Software Engineering (RSE) to denominate this role, as shown in Figure \ref{fig:overviewrobotindustrialprogrammingbpmn}.

The Robotics Software Engineer (RSE) role adds a robot to the workstation after adding and preparing a robot tool if the simulation task must need a tool.  The third step is to set the specific positions in the target trajectory. After, the simulator creates the program using the points created in the previous step. Finally, it is possible to simulate the movement, and the programmer has the visual feedback in the software.  

RSE role validates if the movement is correct. If yes, RSE performs the code generation through the simulators tool and finally sends the final program to the robot that is on the shop-floor. If no, the Robotics Software Engineer rewrites the program. The RSE role acceptance or not the robot behaviour without formal documentation. The validation, in this case, uses the RSE knowledge and expertise about the robot behaviour expected.

Furthermore, the validation process is manual for each program written. It means if there are necessary modifications, the RSE will change the program and validate it again, even if he/she validated part of the program before the modification.The manual aspect for the validation in industrial robot programming increase the project time and cost and decrease industrial robots' reliability. 

\section{Proposed Approach}
\label{sec:PROPOSAL}

There is a widely variety of reliability testing methodologies and one of these is reliability \textbf{acceptance testing} \cite{ELSAYED2012}. Acceptance testing can be a opportunity to address the challenge of writing and design tests for robotics related to the lack of proper channels for coordination and collaboration among multiple teams \cite{AFZAL2020}. Acceptance testing is a bridge between the business and the software product.

But, acceptance testing can be apply manually. In our hypothesis, we decide to apply \textbf{automated testing} to allow that tests will be save and run again after improvements and changes in the code. This is known as regression testing, that is a testing level that guarantee that the modification has not introduce new faults in the code modified \cite{GRAHAM2008}. The benefits of test automation are as follows: increased productivity of testers, better coverage of regression testing, reduced duration of testing phases, reduced cost of software maintenance and increased effectiveness of test cases \cite{SQQA2018}.

Futhermore, we choose \textbf{simulators} as a environment for apply our methodology. Simulator testing plays an important role in complex systems as industrial robots, with simulation is possible to decrease the cost and increase the safety (related to human and system damage) \cite{BOEINGCRASH1,ATKINSON2020}. Our hypothesis is : 
\begin{tcolorbox}[float, colframe=gray!55, coltitle=black, arc=0mm, title=\textbf{Hypothesis}]
\center Automated Acceptance Testing improves reliability for industrial robot program. \end{tcolorbox}

\begin{figure*}[htpb]
\centering
\includegraphics[width=\linewidth]{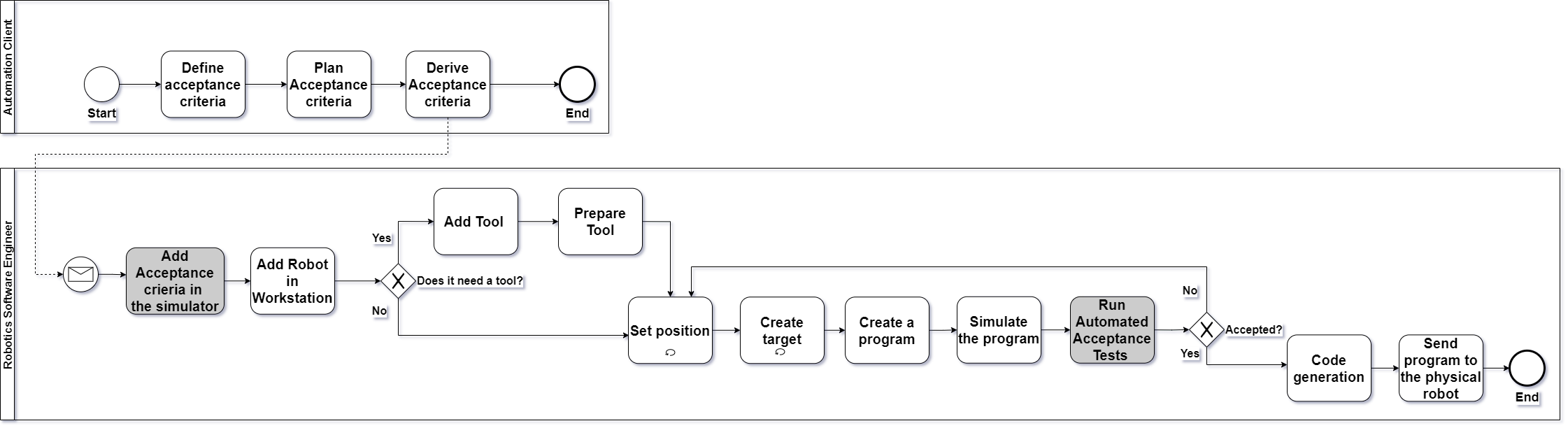}
\caption{Automated Acceptance Testing using Simulators}.
\label{fig:proposal}
\end{figure*}

Figure \ref{fig:proposal} shows how we aim to apply our hypothesis. We put infilled rectangle, the different processes that our proposal will add in the industrial programming using simulators environment. To apply our hypothesis, it will be essential that we have the RSE and a new role, the Automation Client, responsible for defining, planning, and deriving the acceptance criteria. 

After that, the RSE adds the acceptance criteria in the simulator (Figure \ref{fig:proposal}). The simulator performed the followed tasks add robot in workstation, add or not the tool, set positions, create a target, create a program and simulate the program.

At this point, the Automation Client's criteria will be applied to run automated acceptance tests. If the code passes for the acceptance testing, RSE uses the simulator to generate the code and sends it to the physical robot. If no, the Robotics Software Engineer role rewrites the program but uses the acceptance testing results (Figure \ref{fig:proposal}). 

According to the Graham \al \cite{GRAHAM2008}, the followed questions can be part of the reliability acceptance criteria:
(i) What is the current failure rate of the software?, (ii) What will be the failure rate if the customer continues acceptance testing for a long time?, (iii)How many defects are likely to be in the software?, (iv)How much testing has to performed to reach a particular failure rate?.

We aim to adapt these questions for the industrial environment. But it is important to have in mind that our hypothesis uses the simulation testing so the reliability of the system depends on the accuracy of the simulator.

 In this exploration work, we will evaluate the feasibility of the simulator RoboDK. RoboDK is a simulator for industrial robots that provides an intuitive way to program 50 different robot manufacturers. The simulator supports a wide variety of applications, such as pick and place, painting or robot milling. RoboDK API is API for programming industrial robots, that allow industrial robot programming using high level programming language and development environment. Nowadays it is available the following languages: Python, C\#\/.Net, C++ and Matlab \cite{ROBODK2021} .

For the acceptance testing, as our focus is on the testing, we will apply Behavior-Driven Development (BDD) and Acceptance Test-Driven Development (ATDD). Futhermore, we will evaluate the effectiveness of the following frameworks, Jasmine, RSpec and Cucumber, in our proposal \cite{SMART2014, WHYNNE2012, MATSINOPOULOS2020}. 

\section{RELATED WORK}
\label{sec:RELATEDWORK}

Ashraf \al \cite{ASHRAF2020} proposed coverage criteria for white-box testing to test industrial robot tasks and a framework to generate the test cases automatically to achieve the coverage criteria defined by them. Our proposal and their proposal differ from the test level. Our hypothesis is based one acceptance testing, and the authors in the study analyzed are talking about the code coverage criteria (unit, integration or system testing)

In the study performed by Bretl \al \cite{BRETL2008}, the authors presented an algorithm to help to calculate the mass center of legged robots to achieve static equilibrium. The study does not explicitly use software testing techniques, but the authors test the robot's behaviour, which leads us to consider their study as our related work. 

Erich \al \cite{ERICH2019} presented a framework for automatically testing applications for collaborative robots and demonstrated the proposal in a case study for automated testing of a pick and place application. The authors' proposal is a framework applied in a physical environment and at the level of Integration Testing. Our proposal differs from the environment and the test level. We will use Simulator to apply Acceptance Automated Testing to improve the reliability of industrial robots.

In the study performed by Estivill-Castro \al \cite{ESTIVILLCASTRO2018}, the authors proposed the robot simulators follow the Model-View-Controller software patter. They use simulators with the stripped-GUI under a continuous integration paradigm for robots to scale up the testing integration with robot behaviour. Our study and their study are complementary. Both studies aim to use simulators to test behaviours in a robotic environment. The big difference is that our research focuses on a type of robot, which will lead us to solve problems specifics in an industrial environment. 

Mossige \al \cite{MOSSIGE2015}, in their study presented cost-effective automated testing techniques to validate complex industrial robot control systems in an industrial context and employ their methodology in continuous integration and constraint-based testing techniques. The main reason in which our study differs from their study is that the strategy presented by the authors is for validating the timing aspects of distributed systems for complex robot systems. Our study aims to improve the reliability of industrial robots. 

In the study performed by Chung \al \cite{CHUNG2007}, the authors presented a testing process and evaluation items for software testing of intelligent robots. They proposed a test case design methodology based on user requirements and using ISO standards for software testing. The difference between our study and theirs is that our focus is industrial robots. We aim to design and plan and apply automated acceptance testing using simulators to improve industrial robots' reliability. 

\section{CONCLUSIONS}
\label{sec:CONCLUSIONS}

 This article present our hypothesis that automated acceptance testing using simulators improves the reliability of industrial robots.To do it, we performed an overview of industrial robot programming in which we identify the advantages of off-line programming using simulators, and the role play by simulation-testing in testing for robotic system. To create an infrastructure to validate our hypothesis, we have the following future work: (i )evaluate the feasibility of the simulator chosen; (ii) evaluate the feasibility and efficiency of the automated acceptance testing in our proposal; (iii) implementation of solutions to the problem (software prototype); (iv)verification (tests) and validation of the solution ("in vitro"); (v) verification (tests) and validation of the solution ("in vivo") and synthesis and analysis of results.
\bibliographystyle{unsrt}  
\bibliography{references} 

\begin{thebibliography}{10}

\bibitem{IFR2019}
The International~Federation of~Robotics.
\newblock Executive summary world robotics 2019 industrial robots.
\newblock \url{https://ifr.org}, 2020.
\newblock Accessed: 2020-08-25.

\bibitem{FORTUNE2019}
Fortune~Business Insights.
\newblock Industrial robots market size, share \& industry analysis.
\newblock \url{https://www.globenewswire.com}, 2020.
\newblock Accessed: 2020-08-10.

\bibitem{GARCIA2020}
Sergio Garc\'{\i}a, Daniel Str\"{u}ber, Davide Brugali, Thorsten Berger, and
  Patrizio Pelliccione.
\newblock Robotics software engineering: A perspective from the service
  robotics domain.
\newblock In {\em Proceedings of the 28th ACM Joint Meeting on European
  Software Engineering Conference and Symposium on the Foundations of Software
  Engineering}, ESEC/FSE 2020, page 593–604, New York, NY, USA, 2020.
  Association for Computing Machinery.

\bibitem{SQQA2018}
Kshirasagar Naik and Priyadarshi Tripathy.
\newblock {\em Software Testing and Quality Assurance: Theory and Practice}.
\newblock Wiley Publishing, Nova Jersey, EUA, 2nd edition, 2018.

\bibitem{SOMMERVILLE2015}
Ian Sommerville.
\newblock {\em Software Engineering}.
\newblock Addison-Wesley, Boston, Massachusetts, EUA, 2000.

\bibitem{ISO25010}
{{ISO}/{IEC} 25010}.
\newblock {ISO}/{IEC} 25010:2011, systems and software engineering — systems
  and software quality requirements and evaluation (square) — system and
  software quality models.
\newblock Technical report, International Organization for Standardization,
  2011.

\bibitem{ATKINSON2020}
Nancy Atkinson.
\newblock A glitch caused curiosity to freeze in place. but it’s better now.
\newblock \url{https://www.universetoday.com}, 2020.
\newblock Accessed: 2020-09-12.

\bibitem{BOEINGCRASH1}
Gregory Travis.
\newblock How the boeing 737 max disaster looks to a software developer.
\newblock \url{https://spectrum.ieee.org}, 2019.
\newblock Accessed: 2020-09-13.

\bibitem{CHUNG2007}
{Yun Koo Chung} and {Sun-Myung Hwang}.
\newblock Software testing for intelligent robots.
\newblock In {\em 2007 International Conference on Control, Automation and
  Systems}, pages 2344--2349, Seoul, Korea (South), 2007. IEEE.

\bibitem{ASHRAF2020}
Ameena~K Ashraf, Meenakshi D’Souza, and Raoul Jetley.
\newblock Coverage criteria based testing of industrial robots.
\newblock In {\em 2020 IEEE 16th International Conference on Automation Science
  and Engineering (CASE)}, pages 16--21, Hong Kong, China, 2020. IEEE.

\bibitem{BRETL2008}
Timothy Bretl and Sanjay Lall.
\newblock {Testing static equilibrium for legged robots}.
\newblock {\em IEEE Transactions on Robotics}, 24(4):794--807, 2008.

\bibitem{ESTIVILLCASTRO2018}
Vladimir Estivill-Castro, Ren{\'{e}} Hexel, and Carl Lusty.
\newblock {Continuous integration for testing full robotic behaviours in a
  GUI-stripped simulation}.
\newblock {\em CEUR Workshop Proceedings}, 2245:453--464, 2018.

\bibitem{MOSSIGE2015}
Morten Mossige, Arnaud Gotlieb, and Hein Meling.
\newblock {Testing robot controllers using constraint programming and
  continuous integration}.
\newblock {\em Information and Software Technology}, 57(1):169--185, 2015.

\bibitem{ERICH2019}
F.~{Erich}, A.~{Saksena}, G.~{Biggs}, and N.~{Ando}.
\newblock Design and development of a physical integration testing framework
  for robotic manipulators.
\newblock In {\em 2019 IEEE/SICE International Symposium on System Integration
  (SII)}, pages 602--607, Paris, France, 2019. IEEE.

\bibitem{AHMAD2016}
Aakash Ahmad and Muhammad~Ali Babar.
\newblock Software architectures for robotic systems: A systematic mapping
  study.
\newblock {\em Journal of Systems and Software}, 122:16 -- 39, 2016.

\bibitem{TRENDMICRO2021}
Trend Micro's~Forward looking Threat Research (FTR) team in collaboration with
  Politecnico~di Milano~(POLIMI).
\newblock Testing the limits of an industrial robot’s security.
\newblock \url
  {https://www.trendmicro.com/vinfo/us/security/news/internet-of-things/rogue-robots-testing-industrial-robot-security},
  2017 (accessed 12-February-2021).

\bibitem{POGLIANI2020}
Marcello Pogliani, Federico Maggi, Marco Balduzzi, Davide Quarta, and Stefano
  Zanero.
\newblock Detecting insecure code patterns in industrial robot programs.
\newblock In {\em Proceedings of the 15th ACM Asia Conference on Computer and
  Communications Security}, ASIA CCS '20, page 759–771, New York, NY, USA,
  2020. Association for Computing Machinery.

\bibitem{IVALDI2014}
S.~{Ivaldi}, J.~{Peters}, V.~{Padois}, and F.~{Nori}.
\newblock Tools for simulating humanoid robot dynamics: A survey based on user
  feedback.
\newblock In {\em 2014 IEEE-RAS International Conference on Humanoid Robots},
  pages 842--849, Madrid, Spain, 2014. IEEE.

\bibitem{AFZAL2020}
A.~{Afzal}, C.~L. {Goues}, M.~{Hilton}, and C.~S. {Timperley}.
\newblock A study on challenges of testing robotic systems.
\newblock In {\em 2020 IEEE 13th International Conference on Software Testing,
  Validation and Verification (ICST)}, pages 96--107, Porto, Portugal, 2020.
  IEEE.

\bibitem{ROGERS2004}
R.~Owen Rogers.
\newblock Acceptance testing vs. unit testing: A developer's perspective.
\newblock In Carmen Zannier, Hakan Erdogmus, and Lowell Lindstrom, editors,
  {\em Extreme Programming and Agile Methods - XP/Agile Universe 2004}, pages
  22--31, Berlin, Heidelberg, 2004. Springer Berlin Heidelberg.

\bibitem{ELSAYED2012}
Elsayed Elsayed.
\newblock Overview of reliability testing.
\newblock {\em IEEE Transactions on Reliability - TR}, 61:282--291, 06 2012.

\bibitem{GRAHAM2008}
D.~Graham, E.~Van~Veenendaal, and I.~Evans.
\newblock {\em Foundations of Software Testing: ISTQB Certification}.
\newblock Course Technology Cengage Learning, Boston, MA, 2008.

\bibitem{ROBODK2021}
RoboDK.
\newblock Off-line programming.
\newblock \url {https://robodk.com/blog/off-line-programming/}, 2016 (accessed
  28-March-2021).

\bibitem{SMART2014}
J.F. Smart.
\newblock {\em BDD in Action: Behavior-driven development for the whole
  software lifecycle}.
\newblock Manning Publications, Shelter Island, New York, United States, 2014.

\bibitem{WHYNNE2012}
Matt Wynne and Aslak Hellesoy.
\newblock {\em The Cucumber Book: Behaviour-Driven Development for Testers and
  Developers}.
\newblock Pragmatic Bookshelf, Raleigh, North Carolina, United States, 2012.

\bibitem{MATSINOPOULOS2020}
Panos Matsinopoulos.
\newblock {\em Practical Test Automation Learn to Use Jasmine, RSpec, and
  Cucumber Effectively for Your TDD and BDD}.
\newblock Apress, New York, NY, United States, 2020.

\end{thebibliography}
\end{document}